\documentclass{article}

\usepackage{microtype}
\usepackage{graphicx}
\usepackage{subfigure}
\usepackage{booktabs} 

\usepackage{hyperref}

\usepackage[accepted]{icml2023}

\usepackage{amsmath}
\usepackage{amssymb}
\usepackage{mathtools}
\usepackage{amsthm}

\usepackage[capitalize,noabbrev]{cleveref}

\theoremstyle{plain}

\theoremstyle{definition}

\theoremstyle{remark}

\usepackage[textsize=tiny]{todonotes}

\usepackage{pifont} 
\newcommand{\cmark}{\ding{51}}%
\newcommand{\xmark}{\ding{55}}%
\usepackage{arydshln}

\usepackage{xcolor}
\definecolor{noemph}{RGB}{125,125,125}
\newcommand{\noemph}[1]{\textcolor{noemph}{#1}}

\icmltitlerunning{SegCLIP: Patch Aggregation with Learnable Centers for Open-Vocabulary Semantic Segmentation}

\begin{document}

\twocolumn[
\icmltitle{SegCLIP: Patch Aggregation with Learnable Centers for Open-Vocabulary Semantic Segmentation}




\begin{icmlauthorlist}
\icmlauthor{Huaishao Luo}{jdai,swjtu}
\icmlauthor{Junwei Bao}{jdai}
\icmlauthor{Youzheng Wu}{jdai}
\icmlauthor{Xiaodong He}{jdai}
\icmlauthor{Tianrui Li}{swjtu}
\end{icmlauthorlist}

\icmlaffiliation{jdai}{JD AI Research}
\icmlaffiliation{swjtu}{Southwest Jiaotong University, Chengdu, China}

\icmlcorrespondingauthor{Huaishao Luo}{huaishaoluo@gmail.com}
\icmlcorrespondingauthor{Junwei Bao}{baojunwei001@gmail.com}
\icmlcorrespondingauthor{Tianrui Li}{trli@swjtu.edu.cn}

\icmlkeywords{Semantic Segmentation, Open-Vocabulary, ViT, CLIP}

\vskip 0.3in
]



\printAffiliationsAndNotice{}  

\begin{abstract}
Recently, the contrastive language-image pre-training, e.g., CLIP, has demonstrated promising results on various downstream tasks. The pre-trained model can capture enriched visual concepts for images by learning from a large scale of text-image data. However, transferring the learned visual knowledge to open-vocabulary semantic segmentation is still under-explored. In this paper, we propose a CLIP-based model named SegCLIP for the topic of open-vocabulary segmentation in an annotation-free manner. The SegCLIP achieves segmentation based on ViT and the main idea is to gather patches with learnable centers to semantic regions through training on text-image pairs. The gathering operation can dynamically capture the semantic groups, which can be used to generate the final segmentation results. We further propose a reconstruction loss on masked patches and a superpixel-based KL loss with pseudo-labels to enhance the visual representation. Experimental results show that our model achieves comparable or superior segmentation accuracy on the PASCAL VOC 2012 (+0.3\% mIoU), PASCAL Context (+2.3\% mIoU), and COCO (+2.2\% mIoU) compared with baselines. We release the code at \url{https://github.com/ArrowLuo/SegCLIP}. 
\end{abstract} 

\section{Introduction}
\label{sec_introduction}

Semantic segmentation, aiming to assign a label to each pixel of a given image, is an important task and has been researched for a long time. The CNN-based approaches \cite{Long2015FCN, Ronneberger2015UNet, Chen2015Semantic, Zhao2017Pyramid, Deeplabv3plus2018, Wen2022SlotCon} and Transformer-based approaches \cite{Cheng2021Per, Zheng2021SETR, Xie2021SegFormer, Cheng2021mask2former, Jain2022OneFormer} have achieved impressive performance on this topic. However, two significant limitations still need exploration: expensive pixel-level labeling and restricted labeled categories leading to weak generalization \cite{Bucher2019Zero, Xian2019Semantic}. 
\begin{figure}[tp]
    \centering
    \includegraphics[width=0.49\textwidth]{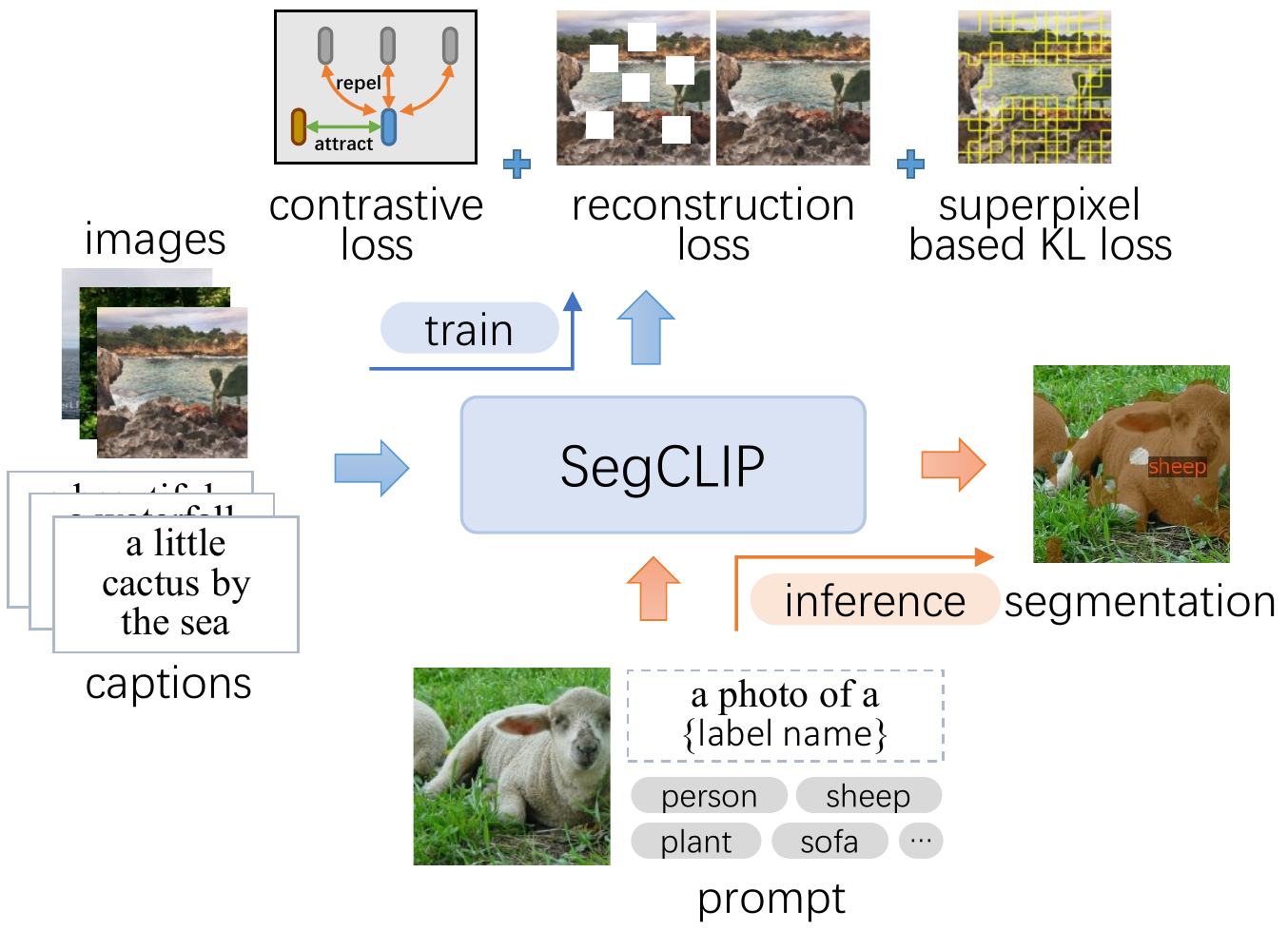}
    \caption{\textbf{Overview of our problem}. The proposed SegCLIP can achieve open-vocabulary semantic segmentation through training with image-text pairs.}
    \label{fig:overview}
\end{figure}

Recent works propose to leverage large-scale image-text pre-trained models to alleviate the above limitations. These works involve zero-shot or weakly supervised semantic segmentation because the large image-text pairs are class-agnostic. Due to the target being to segment an image with arbitrary categories instead of fixed labeling vocabularies, this kind of method is also called open-vocabulary semantic segmentation \cite{Ghiasi2021Scaling,Xu2021A,Liang2022Open,Ma2022Open}. They can be roughly divided into two types. The first is the classification-based method, supervised by the extracted pseudo labels or text features from a pre-trained model, e.g., CLIP \cite{Radford2021Learning}. Moreover, this type is usually achieved with a fully convolutional network or carries out prediction based on mask proposals \cite{zhou2022maskclip,Xu2021A}. The other is to group semantic regions along training with large-scale image-text datasets, which can be called the group-based method \cite{Xu2022GroupViT}. Through different routes, the fundamental logic behind them is that the image-text pre-trained model can learn vision-text alignment from image-level to pixel-level features. Some interpretability methods, like CAM \cite{selvaraju2017grad} and Transformer-interpretability \cite{chefer2021transformer}, can support such an argument, such as in the work of \cite{Zabari2021Semantic}.

Following the research line of learning pixel-level alignment from image-text pairs, we explore the semantic regions with the group-based method in this paper. Compared with the classification-based method, which involves mask proposals and label classification, the group-based method is straightforward. It has consistent objectives with the pretraining model, e.g., training with a contrastive loss using image-text pairs. Further, the group-based model jointly learns visual and textual representations as humans do, so it has the potential to be improved from a multimodal perspective. Instead of training from scratch, the group-based method can also benefit from the pre-trained model.

To this end, we propose a group-based model SegCLIP to accomplish open-vocabulary semantic segmentation. The SegCLIP can be regarded as \textit{segmentation}+CLIP. Specifically, the proposed model has a similar architecture to the CLIP but a modified image encoder. The image encoder is based on the ViT (Vision Transformer) \cite{Dosovitskiy2021An}. Instead of operating on regular grids, we designed a plugged semantic group module to aggregate patches with learnable centers. The learnable centers can dynamically merge visual patches to semantic concepts via a mapping matrix generated by a cross-attention mechanism. This plugged group module can be inserted into the middle layers of the image encoder to generate irregular-shaped segments. Thus, the SegCLIP can transfer knowledge from CLIP to semantic segmentation. We use a small number of image-text pairs to train our experiments' extra randomly initialized parameters. Figure \ref{fig:overview} illustrates the training and inference process. During inference, the label name is filled into a given prompt format, and the semantic segments are obtained by calculating the similarity between the text representation and the semantic groups.

Moreover, we propose two auxiliary losses to enhance the visual representation for semantic segmentation. One is a reconstruction loss, which aims to recover the masked patches through their visual context. Such a reconstruction loss is effective from the previous work \cite{He2022Masked,wang2022bevt,Zhou2022iBOT}. The difference is that our reconstruction process is designed based on irregular-shaped segments with a mapping matrix instead of regular patches. The other is a KL loss (Kullback-Leibler divergence Loss) used to learn a better mapping matrix via the superpixel label, which can be obtained via the off-the-shelf tool. The KL loss can keep the consistency of pixel-level features.

\section{Model}
\label{sec_model}
\begin{figure*}[htbp] 
    \centering
    \includegraphics[width=0.98\textwidth]{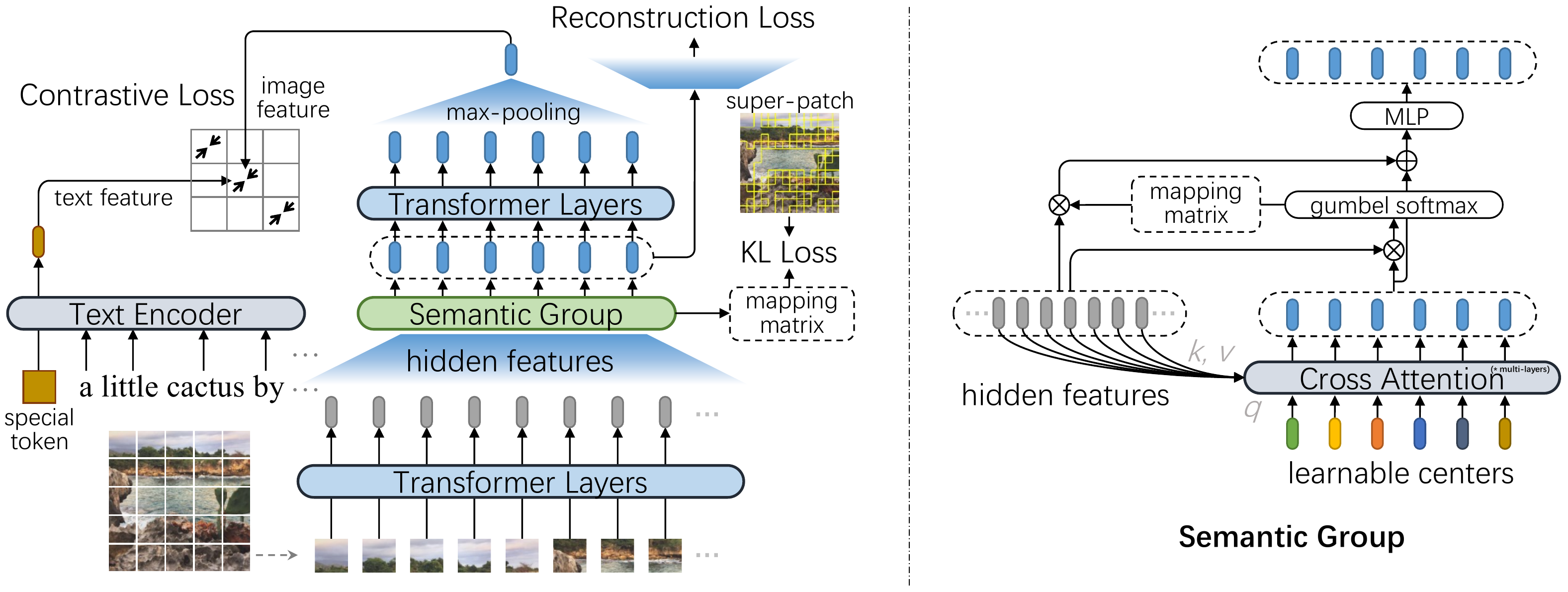}
    \caption{\textbf{The framework of SegCLIP}. The SegCLIP is a dual-encoder architecture containing a text and image encoder. The semantic group module (zoom in at the right) is proposed to generate regular patches to arbitrary-shaped semantic regions. Three losses, including contrastive loss, reconstruction loss, and superpixel-based KL loss, are used in training.}
    \label{fig:main_structure}
\end{figure*}

Figure \ref{fig:main_structure} presents the SegCLIP as a dual-encoder architecture. One encoder is for text representation, and the other is for image representation. We propose a plugged semantic group module to aggregate patches with learnable centers in the image encoder, thus injecting the CLIP with the capacity to deal with semantic segmentation. The backbone of SegCLIP is the ViT version of CLIP, and the details can be found in \cite{Radford2021Learning}. We describe the architecture of SegCLIP, training losses, and inference process in detail in this section.

\subsection{Main Architecture}
The architecture of SegCLIP mainly consists of a text encoder $\mathcal{E}_T(\cdot)$ and an image encoder $\mathcal{E}_I(\cdot)$, similar to the CLIP. Such a design can transfer the knowledge naturally from the pre-trained weights of the CLIP instead of training from scratch. Nevertheless, it takes work to achieve semantic segmentation directly because the CLIP is pre-trained with image-level features and needs help to finish pixel-level tasks. We propose a plugged semantic group module within the image encoder with learnable centers to aggregate the low-layer pixel features to achieve the segmentation. The learnable centers can be regarded as semantic regions and gather semantical pixels along with the training process. Thus, the SegCLIP can finish open-vocabulary semantic segmentation.

As shown in Figure \ref{fig:main_structure}, the model's input is a pair of text $\mathbf{T}=\{w_i\}_{i=1}^{M}$ and image $\mathbf{I}=\{p_j\}_{j=1}^{N}$, where $w_i$ means the $i$-th token within the text, $p_j$ means the $j$-th non-overlapped patches of the image, $M$ and $N$ denotes the number of given text and image, respectively. Following the ViT version of CLIP, the token is generated via a lower-cased byte pair encoding (BPE), and the tokens representation $\{\mathbf{e}_{w_i}\}_{i=1}^{M}$ and patches representation $\{\mathbf{e}_{p_j}\}_{j=1}^{N}$ are obtained by an Embedding operation and a linear projection, respectively. Then the tokens representation is fed into Transformer layers \cite{Vaswani2017} to generate the final text feature as $\mathbf{z}_w = \mathcal{E}_T\big(\{\mathbf{e}_{w_i}\}_{i=1}^{M}\big)$. The image representation is fed into other Transformer layers plus the semantic group module to generate the final image feature as $\mathbf{z}_p = \mathcal{E}_I\big(\{\mathbf{e}_{p_j}\}_{j=1}^{N}\big)$. Finally, the contrastive loss can be calculated on the text feature $\mathbf{z}_w$ and the image feature $\mathbf{z}_p$. In our setting, the text feature $\mathbf{z}_w$ comes from a special token $\mathtt{[SEP]}$, which is appended as the last token of the text. The image feature $\mathbf{z}_p$ is generated by the last Transformer layer followed by a $\texttt{max-pooling}$ operation.

\subsection{Semantic Group Module}
To gather the regular patches to arbitrary-shaped semantic regions, we design a semantic group to plug into the Transformer layers of the image encoder. In other words, the semantic group module can be regarded as the second stage of the image encoder, with different Transformer layers as the first and third stages. Assuming the patches representation is $\mathcal{H}_p=\{\mathbf{h}_{p_j}^s\}_{j=1}^{N}$ after passing through the first stage's $s$-th (also the last) Transformer layer. The semantic group module can gather different patches by calculating semantic similarity. Specifically, we first randomly initialize a group of learnable centers $\mathcal{H}_c=\{\mathbf{c}_k\}_{k=1}^L$, then obtain contextual centers $\hat{\mathcal{H}}_c=\{\hat{\mathbf{c}}_k\}_{k=1}^L$ through some cross-attention layers as follows,
\begin{align}
	\hat{\mathcal{H}}_c^t = \texttt{CrossAttention}(\mathcal{H}_c^t, \mathcal{H}_p, \mathcal{H}_p),
\end{align}
where $t$ is the layer number of cross-attention, the start $\mathcal{H}_c^1$ is the $\mathcal{H}_c$, and the $\hat{\mathcal{H}}_c$ is the last $\hat{\mathcal{H}}_c^t$, the $\texttt{CrossAttention}$ is a cross-attention layer, the same as the Self-Attention layer in Transformer \cite{Vaswani2017}, but the input is asymmetrically separate embedding sequences, in here, the query is $\mathcal{H}_c^t$, and the key and value are $\mathcal{H}_p$, respectively.

After obtaining the contextual centers $\hat{\mathcal{H}}_c$, we can assign each patch to a corresponding center via a mapping matrix $\mathcal{M} \in \mathbb{R}^{N \times L}$ generated by the $\texttt{Gumbel-Softmax}$ operation \cite{Jang2017Gumbel,Xu2022GroupViT}. 
\begin{align}
	\mathcal{M} = \texttt{Gumbel-Softmax}(\mathcal{H}_p\hat{\mathcal{H}}_c^{\top}),
\end{align}
where each row of $\mathcal{M}$ is a one-hot vector, and $\mathcal{M}_{jk}$ denotes the $j$-th patch belongs to $k$-th semantic center if its value is 1. The $\mathcal{M}$ keeps a patch belonging to only and if only a center, which benefits the final semantic segmentation. 

Finally, we can calculate the representation of semantic regions $\hat{\mathcal{H}}_p$ with the patches representation $\mathcal{H}_p$, the mapping matrix $\mathcal{M}$, and the contextual centers $\hat{\mathcal{H}}_c$ as follows,
\begin{align}
	\hat{\mathcal{H}}_p = \texttt{MLP}\big(\texttt{MEAN}(\mathcal{M}^{\top}\mathcal{H}_p) + \hat{\mathcal{H}}_c \big),
\end{align}
where $\texttt{MEAN}$ denotes doing the average for each center using the patches belonging to it. $\texttt{MLP}$ is a multilayer perceptron block containing two fully-connected layers and a GELU \cite{hendrycks2016gaussian} between them.

The generated representation of semantic regions $\hat{\mathcal{H}}_p$ is fed to the Transformer layers of the third stage to learn sufficiently interactive region features $\mathcal{Z}_p$ further.

\subsection{Reconstruction Loss}
\begin{figure}[tp]
    \centering
    \includegraphics[width=0.40\textwidth]{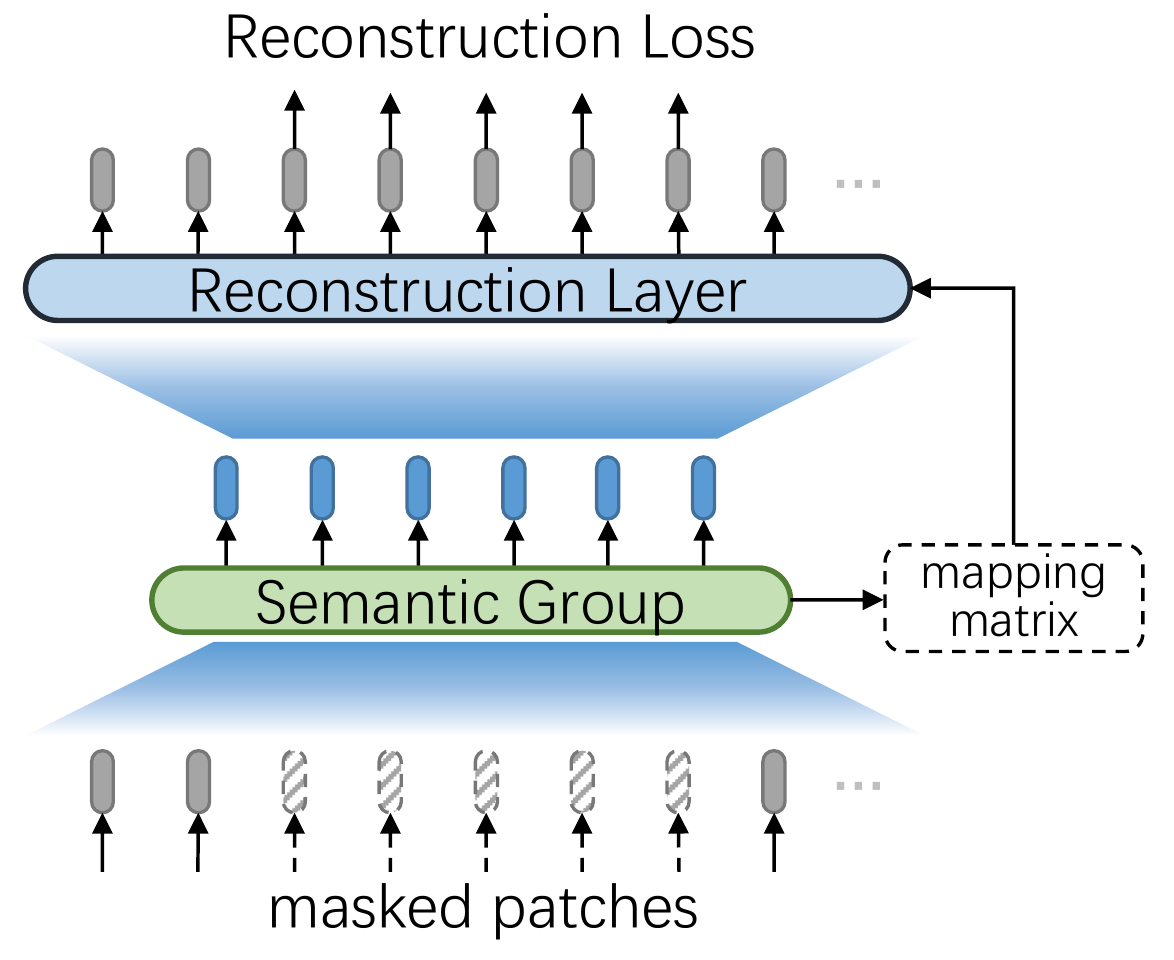}
    \caption{\textbf{Reconstruction Loss}.}
    \label{fig:reconstruction_loss}
\end{figure}

In addition to the contrastive loss, we propose a self-supervised reconstruction loss to enhance the visual representation for segmentation. As shown in Figure \ref{fig:reconstruction_loss}, the reconstruction loss aims to recover the masked patches through their visual context, similar to MAE \cite{He2022Masked}. The difference is that our reconstruction process is designed based on irregular-shaped segments with a mapping matrix. 

We first generate a masked version of region representation $\hat{\mathcal{H}}_p^{(m)}$ and mapping matrix $\mathcal{M}^{(m)}$ via the semantic group module on the unmasked patches for the MAE encoder. However, the region representation can not be used to calculate the reconstruction loss because the unmasked patches have been gathered into different regions. We propose a reconstruction layer to restore the representation of patches from $\hat{\mathcal{H}}_p^{(m)}$ as,
\begin{align}
	\tilde{\mathcal{H}}_p^{(m)} = \text{GELU}\big(\texttt{Linear}(\mathcal{M}^{(m)})^{\top}\hat{\mathcal{H}}_p^{(m)}\big),
\end{align}
where \texttt{Linear} is a fully-connected layer, and the \text{GELU} is the activation function.
Then we use extra Transformer layers, similar to the third stage of the image encoder, to obtain the final representation $\mathcal{Z}_p^{(m)}$ using the $\tilde{\mathcal{H}}_p^{(m)}$.

We keep the MAE decoder as in \cite{He2022Masked} with the input $\mathcal{Z}_p^{(m)}$. Finally, the reconstruction loss is the mean squared error (MSE) between the reconstructed image $\mathbf{I}^{(m)}$ and the original image $\mathbf{I}$, $\mathcal{L}_{rec} = \texttt{MSE}(\mathbf{I}^{(m)}, \mathbf{I})$. 

\subsection{Superpixel based KL Loss}
\begin{figure}[tp]
    \centering
    \includegraphics[width=0.35\textwidth]{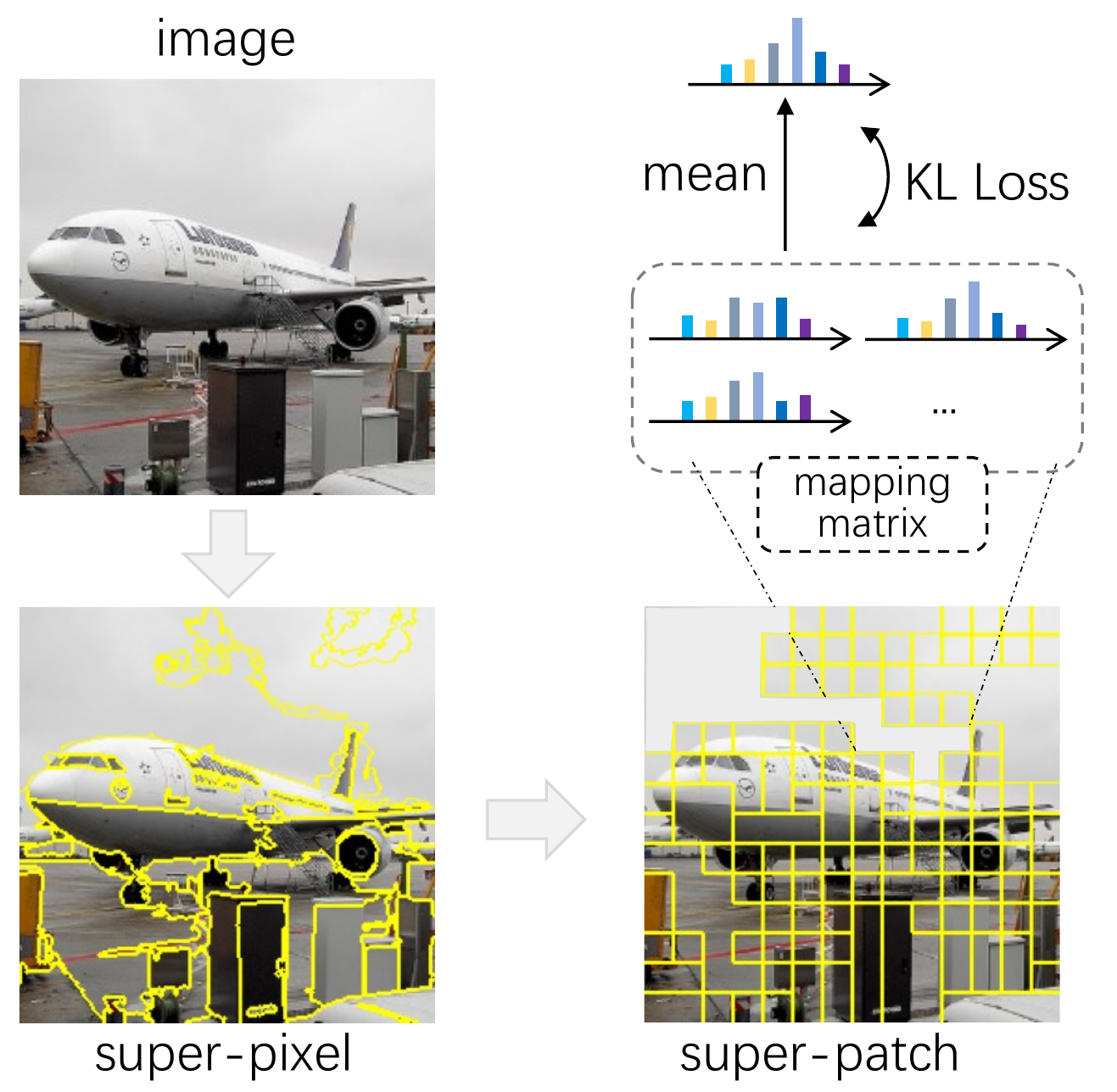}
    \caption{\textbf{Superpixel based KL Loss}.}
    \label{fig:klloss}
\end{figure}

Besides the reconstruction loss, we propose a superpixel-based KL loss to guild the learning of a mapping matrix. The motivation is to keep the pixel-level consistency when gathering the patches to regions. Intuitively, the pixels of a superpixel should be gathered into a region instead of one more region. The calculation process is illustrated in Figure \ref{fig:klloss}. For a given image, we first obtain its superpixel with a graph-based segmentation method from \cite{felzenszwalb2004efficient}, which is unsupervised and does not need to train on any datasets. There are many other superpixel methods, but we chose this typical one as a demonstration. 

Assuming there are some superpixels, each pixel in the same superpixels has the same label, e.g., superpixel id. Thus for each patch, we assign it a label, e.g., the average floor value of ids from its pixels. Thus, we can obtain a super-patch corresponding to the superpixel. Intuitively, the patches within a super-patch are also covered by a superpixel. Note that a superpixel id is a number used to distinguish different superpixels, and we do not care about its meaning in the loss calculation. Every patch of a super-patch should have a consistent probability in the mapping matrix $\mathcal{M}$ because they should be gathered in a region. In other words, the probability of a patch in the mapping matrix should be similar to the average probability of the patches within the same super-patch. Thus, a symmetric KL loss is designed as follows,
\begin{align}
    &\hat{\mathcal{P}}_j = \texttt{softmax} \big( \frac{1}{|\mathcal{G}_j|} \sum_{\hat{j} \in \mathcal{G}_j} \mathcal{P}_{\hat{j}} \big), \\
    &\mathcal{L}_{sup} = \frac{1}{2N} \! \sum_{j=1}^{N} \! \big( \texttt{KL}(\mathcal{P}_j, \! \hat{\mathcal{P}}_j) \!+\! \texttt{KL}(\hat{\mathcal{P}}_j, \! \mathcal{P}_j)\big),
\end{align}
where $\texttt{KL}$ is the Kullback-Leibler divergence, $\mathcal{P}_j$ is the regions' probability of $j$-th patch, which is obtained by the $j$-th row of $\mathcal{M}$ after $\texttt{softmax}$ operation, and $\mathcal{G}_j$ is the indexes of the patches contained in a super-patch which also contains the $j$-th patch. By decreasing the $\mathcal{L}_{sup}$, the model tends to gather the patches within a superpixel together, which benefits the segmentation.

\subsection{Training and Inference}
\paragraph{Training}
In addition to the reconstruction loss $\mathcal{L}_{rec}$ and the superpixel-based KL loss $\mathcal{L}_{sup}$, the model is also trained with the contrastive loss $\mathcal{L}_{con}$ in an end-to-end manner. The total loss is the sum of them,
\begin{align}
    \mathcal{L}_{total} = \mathcal{L}_{con} + \mathcal{L}_{rec} + \mathcal{L}_{sup}.
\end{align}

The $\mathcal{L}_{con}$ is a symmetric cross-entropy loss calculated on the text feature $\mathbf{z}_w$ and image feature $\mathbf{z}_p$, similar to the CLIP \cite{Radford2021Learning},
\begin{align}
    \mathcal{L}_{con} &= \frac{1}{2} (\mathcal{L}_{t2i} + \mathcal{L}_{i2t}), \\
    \mathcal{L}_{t2i} &= \!\! -\frac{1}{\mathcal{B}} \!\! \sum_i^{\mathcal{B}} \! {\log \frac{\exp \big(s(\mathbf{z}_w^{(i)}, \mathbf{z}_p^{(i)}) \big)}{\sum_{j=1}^{\mathcal{B}}{\exp \big(s(\mathbf{z}_w^{(j)}, \mathbf{z}_p^{(i)} \big)}}}, \label{eq:t2i} \\
    \mathcal{L}_{i2t} &= \!\! -\frac{1}{\mathcal{B}} \!\! \sum_i^{\mathcal{B}} \! {\log \frac{\exp \big(s(\mathbf{z}_w^{(i)}, \mathbf{z}_p^{(i)}) \big)}{\sum_{j=1}^{\mathcal{B}}{\exp \big(s(\mathbf{z}_w^{(i)}, \mathbf{z}_p^{(j)} \big)}}}, \label{eq:i2t}
\end{align}
where, $s(\mathbf{z}_i, \mathbf{z}_j) = \frac{\mathbf{z}_i \mathbf{z}_j^\top}{\lVert \mathbf{z}_i \rVert \lVert \mathbf{z}_j \rVert}$ is the cosine similarity, $\mathcal{B}$ is the batch size, the superscripts $(i)$ and $(j)$ means the $i$-th and $j$-th sample, respectively.

\paragraph{Inference} Due to the learned mapping matrix, the SegCLIP can finish semantic segmentation without further finetuning on any datasets. The image feature can be obtained through the image encoder for a segmentation task with candidate labels. For the text feature, we use a template $\texttt{a photo of a \{label name\}.}$ to form the input text of the text encoder with different label names. Specifically, we use the $\mathcal{Z}_p \in \mathbb{R}^{L \times H}$ from the last Transformer layer of the image encoder as the representation of each region, where $L$ is the number of learnable centers, and $H$ is the hidden size. The label features can be denoted as $\{\mathbf{z}_w^{(\tau)}\}_{\tau=1}^{\mathcal{T}}$ if there are $\mathcal{T}$ candidate labels. After calculating the cosine similarity of each row of $\mathcal{Z}_p$ and each $\mathbf{z}_w^{(\tau)}$ via $s(\mathbf{z}_i, \mathbf{z}_j)$ from Eqs. (\ref{eq:t2i}-\ref{eq:i2t}), we can obtain a similarity matrix $\hat{\mathcal{S}} \in \mathbb{R}^{L \times \mathcal{T}}$, in which each row denotes the labels' probability of a region. Then the similarity $\mathcal{S} \in \mathbb{R}^{N \times \mathcal{T}}$ between patches and candidate labels can be calculated using the mapping matrix $\mathcal{M} \in \mathbb{R}^{N \times L}$ via $\mathcal{S} = \mathcal{M}\hat{\mathcal{S}}$. We can assign each patch a label with the highest similarity of each row from the $\mathcal{S}$. We can further execute an interpolation operation on the $\mathcal{S}$ from $N$ to image size to obtain a pixel-level assignment matrix, then an irregular-shaped and pixel-level segmentation.

\section{Experiments}
\label{sec_experiments}
We first describe datasets and implementation details before ablating various settings of our model. Then we present the state-of-the-art results on three datasets in an annotation-free manner. Finally, we demonstrate some qualitative results of our model.

\subsection{Datasets}
We pretrain the SegCLIP on the training splits of Conceptual Captions (CC) \cite{Sharma2018CC} and COCO \cite{Lin2014COCO}, which contain 3M and 400K image-text pairs, respectively. 

For the semantic segmentation, we evaluate the model on the validation splits of the PASCAL VOC 2012 \cite{Everingham2010VOC}, PASCAL Context \cite{Mottaghi2014Context}, and COCO datasets. These datasets contain  20, 59, and 80 foreground classes, respectively. To distinguish the foreground classes from the background, we set the threshold to 0.75, 0.25, and 0.65 on the similarities for PASCAL VOC 2012, PASCAL Context, and COCO, respectively. The metric is the mIoU calculated with the predicted and ground truth segmentation masks. The short side of the given image is resized to 224 during inference.

\subsection{Experimental Details}
The architecture is based on the ViT version of CLIP, and the text encoder and image encoder are all 12 Transformer layers. The image size is set to 224 $\times$ 224, and the patch size is 16 $\times$ 16. The max length of the text tokens is 32. We initialize the embedding and Transformer layers from the CLIP pre-trained weight as default. For the semantic group module, we put it after the 10th Transformer layer in the image encoder via grid search based on segmentation datasets. The cross-attention layer number is set to 2. The decoder layer of MAE is 3, and the mask rate of patches is 0.75. The number of learnable centers is 8. We randomly initialize the parameters of the semantic group module, MAE decoder, and the rest of the Linears in the model. For the optimization, we use Adam optimizer and a cosine schedule of learning rate following the CLIP. The initial learning rate is 4e-6 for the embedding layers, text encoder, and Transformer layers of the image encoder before the semantic group module. For the rest of the parameters, the initial learning rate is 4e-3. We pretrain our model using 8 NVIDIA A100 GPUs with a batch size of 768 for 10 epochs. This process takes approximately 6 hours.

\subsection{Ablation Studies}
We conduct ablation studies on the designed losses and key hyperparameters to introduce their influence in this section.

\vspace{0.1cm}
\noindent
\textbf{Effective of the Reconstruction Loss} \quad
 As shown in Table \ref{tab:result_of_ablation}, training with the reconstruction loss can improve the mIoU by 1.19\%, 0.92\%, and 0.66\% on PASCAL VOC, PASCAL Context, and COCO, respectively under the condition of without the superpixel-based KL loss and can improve the mIoU by 4.11\%, 0.56\%, and 0.52\% under the condition of with the superpixel based KL loss. The results demonstrate that the well-designed reconstruction loss restoring the masked patches with contextual visual features can enhance the encoder and make the mapping matrix a better match for the segmentation task.

 \vspace{0.1cm}
 \noindent
 \textbf{Effective of the Superpixel based KL Loss} \quad
We also report the consistent improvement of the superpixel-based KL loss in Table \ref{tab:result_of_ablation}. The gains are 0.54\%, 0.72\%, and 1.07\%, and 3.46\%, 0.36\%, and 0.93\% on PASCAL VOC, PASCAL Context, and COCO under the condition of without or with the reconstructing loss, respectively. We suppose that the pseudo superpixel labels can keep the pixel-level visual feature relatively consistent within the segments. 
\begin{table}[tp]
    \setlength{\tabcolsep}{5pt}
    \centering
    \begin{tabular}{cccccc}
        \toprule
        Model       & R-Loss & S-KL & VOC & Context & COCO \\ \midrule
        SegCLIP     &   &   & 47.95 & 23.43 & 24.86 \\
        SegCLIP     & \cmark &   & 49.14 & 24.35 & 25.52 \\
        SegCLIP     &   & \cmark  & 48.49 & 24.15 & 25.93 \\
        SegCLIP     & \cmark & \cmark & \textbf{52.60} & \textbf{24.71} & \textbf{26.45} \\
        \bottomrule
    \end{tabular}
    \caption{\textbf{Ablation of the proposed losses (mIoU)}. \emph{R-Loss} is the reconstruction loss, \emph{S-KL} is the superpixel-based KL loss.}
    \label{tab:result_of_ablation}
\end{table}
\begin{table}[tp]
    \setlength{\tabcolsep}{5pt}
    \centering
    \begin{tabular}{cccccc}
        \toprule
        Model     & P-Ly & C-NO. & VOC & Context & COCO \\ \midrule
        SegCLIP     & 6  & 8 & 35.28 & 19.28 & 16.73 \\
        SegCLIP     & 8  & 8 & 43.75 & 22.71 & 21.40 \\
        SegCLIP     & 10  & 6 & 47.03 & 23.36 & 24.85 \\
        SegCLIP     & 10  & 8 & \textbf{47.95} & 23.43 & \textbf{24.86} \\
        SegCLIP     & 10  & 10 & 44.89 & \textbf{23.46} & 24.74 \\
        SegCLIP     & 11  & 8 & 22.07 & 10.76 & 12.08 \\
        \bottomrule
    \end{tabular}
    \caption{\textbf{Ablation of plugged layer (P-Ly) and center number (C-NO.) of semantic group module}. The results are obtained with only the contrastive loss.}
    \label{tab:result_of_plug_group}
\end{table}

\vspace{0.1cm}
\noindent
\textbf{Influence of the Plugged Layer} \quad 
In Table \ref{tab:result_of_plug_group}, we conduct experiments on different plugged layers of the semantic group module, from 6 to 11, with the same 8 learnable centers. The results reflect that the plugged layer 10 can achieve better performance than other plugged points, and too small and big numbers decrease the mIoU significantly. We suppose that too small plugged points may harm the pre-trained CLIP weights, and the low-layer feature is segments-irrelevant. The features from big plugged points are also segments-irrelevant and can not benefit the segmentation task.

\vspace{0.1cm}
\noindent
\textbf{Influence of the Center Number} \quad
We also conduct experiments on different learnable centers of the semantic group module with the same plugged layer 10. The results in Table \ref{tab:result_of_plug_group} demonstrate that the 8 learnable centers can achieve better or comparable performance. The mIoU is not sensitive on 6, 8, and 10 learnable centers. We chose 8 as the default hyperparameter in this work.

\vspace{0.1cm}
\noindent
\textbf{Influence of the Cross-Attention Layer} \quad
Table \ref{tab:result_of_group_layer} shows the results on different layers of the cross-attention layer in the semantic group module. Compared with the mIoU obtained by training without a cross-attention layer, training with a cross-attention layer can achieve better performance. Such a phenomenon suggests that the cross-attention layer can make the learnable centers match better with the features of patches and focus on different parts of the given image. We also obverse that two cross-attention layers achieve better mIoU than others, but the results from numbers 1 and 3 are comparable. A large number of cross-attention layers, e.g., 4, may harm the performance. We consider that our training datasets are insufficient to train deep layers.
\begin{table}[tp]
    \setlength{\tabcolsep}{6pt}
    \centering
    \begin{tabular}{cccccc}
        \toprule
        Model     & Cross-Att. & VOC & Context & COCO \\ \midrule
        SegCLIP     & 0 & 44.44 & 22.28 & 22.11 \\
        SegCLIP     & 1 & 47.63 & 23.29 & 24.21 \\
        SegCLIP     & 2 & \textbf{47.95} & \textbf{23.43} & \textbf{24.86} \\
        SegCLIP     & 3 & 47.83 & 23.24 & 24.70 \\
        SegCLIP     & 4 & 45.39 & 23.17 & 23.80 \\
        \bottomrule
    \end{tabular}
    \caption{\textbf{Ablation of cross-attention layer (Cross-Att.)} The plugged layer is 10, and the center NO. is 8. The results are obtained with only the contrastive loss.}
    \label{tab:result_of_group_layer}
\end{table}
\begin{table*}[tp]
    \setlength{\tabcolsep}{6pt}
    \centering
    \begin{tabular}{cccccccccc}
        \toprule
        Model       & Arch.  & Init.  & Training Data & Sup. & Zero-Shot & VOC & Context & COCO \\ \midrule 
        \noemph{DeiT$^\natural$ \cite{Touvron2021Training}}    &  \noemph{ViT}      &        & \noemph{ImageNet} & \noemph{Class} & \noemph{\xmark} & \noemph{53.0} & \noemph{35.9} & \noemph{-} \\
        \noemph{DINO$^\natural$ \cite{Caron2021Emerging}}    &  \noemph{ViT}      &        & \noemph{CC12M+YFCC} & \noemph{Self} & \noemph{\xmark} & \noemph{37.6} & \noemph{22.8} & \noemph{-} \\
        \noemph{MoCo$^\natural$ \cite{Chen2021An}}   &  \noemph{ViT}      &        & \noemph{CC12M+YFCC} & \noemph{Self} & \noemph{\xmark} & \noemph{36.1} & \noemph{23.0} & \noemph{-} \\
        GroupViT \cite{Xu2022GroupViT}    &  ViT      &        & CC12M+YFCC & Text & \cmark & 52.3 & 22.4 & 24.3 \\  \cdashline{1-9}[3pt/4pt]
        GroupViT$_{\text{1-s}}$   &  ViT      &        & CC+COCO & Text & \cmark & 28.1 & 14.8 & 12.9 \\
        GroupViT$_{\text{2-s}}$   &  ViT      &        & CC+COCO & Text & \cmark & 19.7 & 10.4 & 8.0 \\
        SegCLIP (ours)     &  ViT      &        & CC+COCO & Text & \cmark & 33.3 & 19.1 & 15.2 \\ 
        SegCLIP (ours)     &  ViT      & \cmark & CC+COCO & Text & \cmark & \textbf{52.6} & \textbf{24.7} & \textbf{26.5} \\
        \bottomrule
    \end{tabular}
    \caption{\textbf{Comparison of different models on mIoU}. `Arch.' and `Sup.' are short for architecture and supervision, respectively. `Init.' means whether be initialized with CLIP. CC12M and YFCC are from \cite{Changpinyo2021CC12M} and \cite{Thomee2016YFCC100M}, respectively. $^\natural$ means results from \cite{Xu2022GroupViT}. GroupViT$_{\text{1-s}}$ and GroupViT$_{\text{2-s}}$ are our implementations on the CC and COCO datasets, with one-stage and two-stage grouping blocks, respectively.}
    \label{tab:result_of_comparison} 
\end{table*}

\subsection{Comparisons with State-of-the-Art Methods}
As shown in Table \ref{tab:result_of_comparison}, we compare our model against class-supervised, visually self-supervised, and textually supervised baselines. The results of class-supervised and visually self-supervised baselines are obtained from \cite{Xu2022GroupViT}. They are pixel-wise classification models finetuned on the pre-trained ViT models, i.e., DeiT \cite{Touvron2021Training}, DINO \cite{Caron2021Emerging}, and MoCo \cite{Chen2021An}, with a 1$\times$1 convolutional layer as the semantic segmentation head. The finetuning datasets are the training sets of the VOC and Context separately. Compared with the class-supervised model, our result (52.5\%) on VOC is still comparable (53.0\%), although training without manually pixel-level annotations. 
\begin{figure}[tp]
    \centering
    \includegraphics[width=0.46\textwidth]{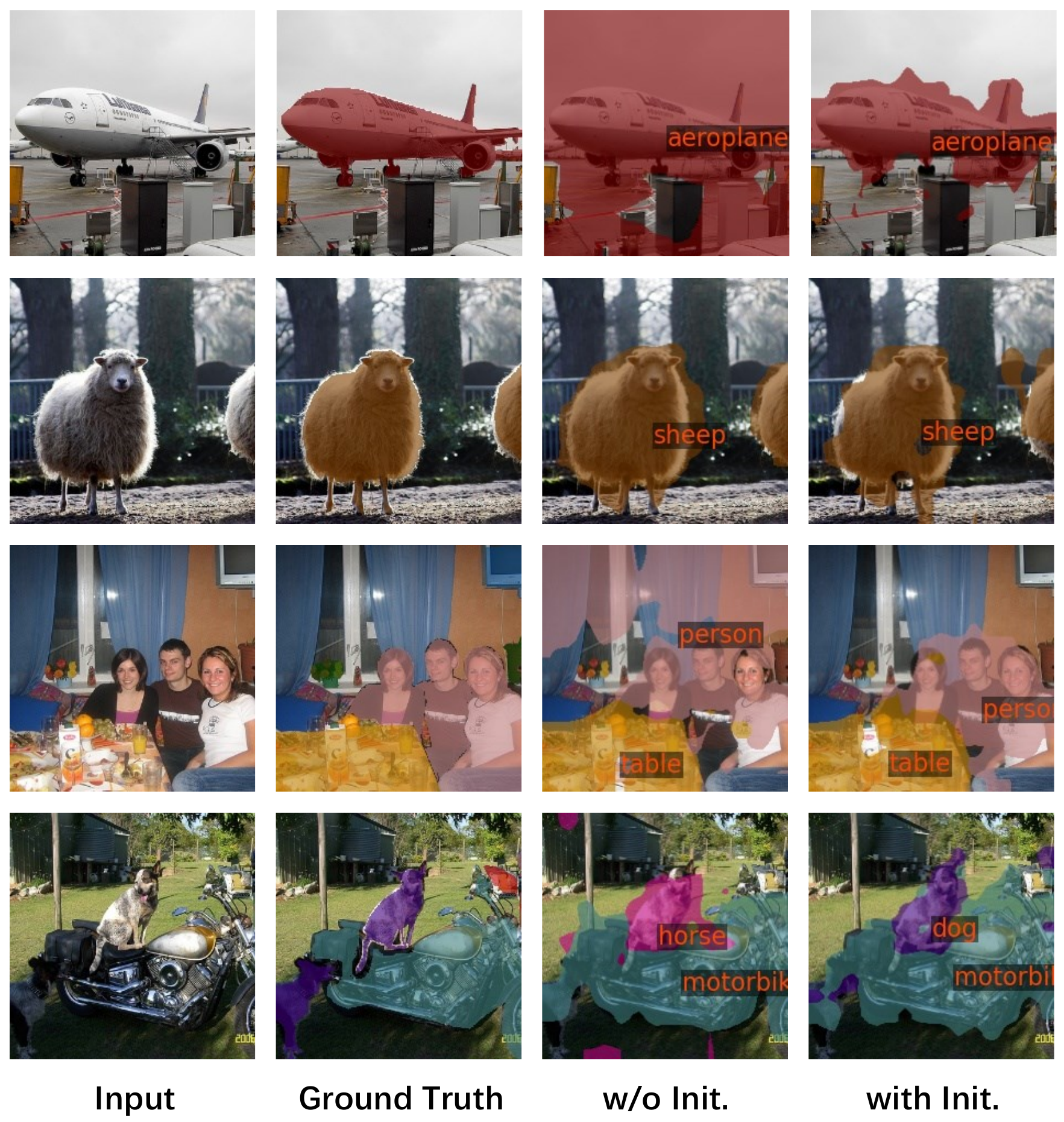}
    \caption{\textbf{Qualitative results on PASCAL VOC}.}
    \label{fig:qualitative_results_voc}
\end{figure}
\begin{figure}[tp]
    \centering 
    \includegraphics[width=0.46\textwidth]{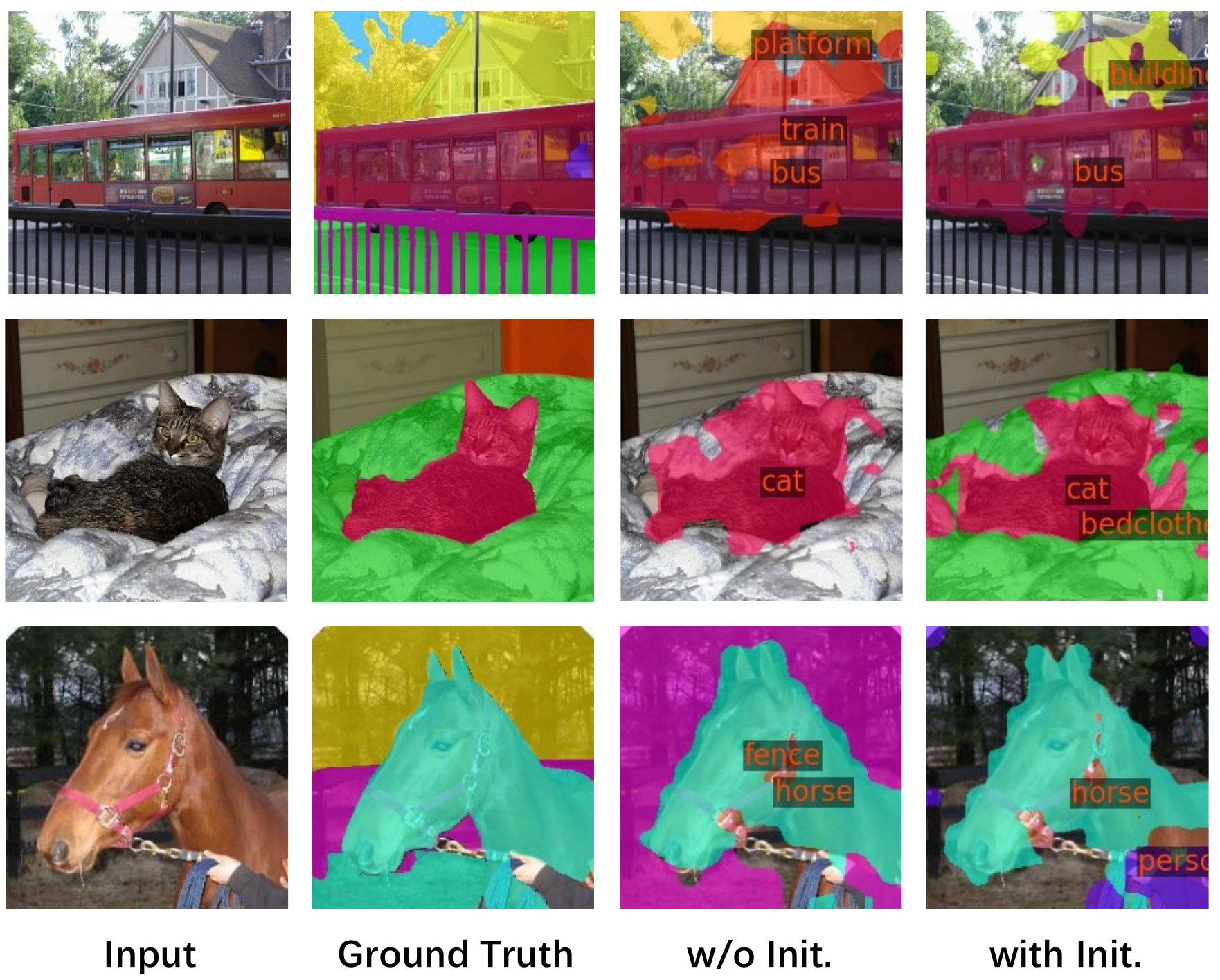}
    \caption{\textbf{Qualitative results on PASCAL Context}.}
    \label{fig:qualitative_results_context}
\end{figure}

Compared with the state-of-the-art textually supervised method GroupViT, our initialized SegCLIP achieves 0.3\%, 2.3\%, and 2.2\% gains on the VOC, Context, and COCO, respectively. We also conduct experiments for GroupViT on CC and COCO datasets for a fair comparison. Our SegCLIP trained from scratch achieves 5.2\%, 4.3\%, and 2.3\% improvements compared with the GroupViT$_{\text{1-s}}$. Note that the GroupViT$_{\text{1-s}}$ achieves superior accuracy than GroupViT$_{\text{2-s}}$ in our settings. We suppose the CC and COCO, which are smaller than the CC12M \cite{Changpinyo2021CC12M} and YFCC \cite{Thomee2016YFCC100M}, may lead the unstable and insufficient training for the 2-stage GroupViT. When initialized with the pre-trained CLIP, the SegCLIP improves the mIoU by 19.3\%, 5.6\%, and 11.3\% on the VOC, Context, and COCO compared with training from scratch, respectively. It implies that our model could benefit from the pre-trained CLIP, which also proves the flexibility of the semantic ground module.
\begin{figure}[tp]
    \centering
    \includegraphics[width=0.46\textwidth]{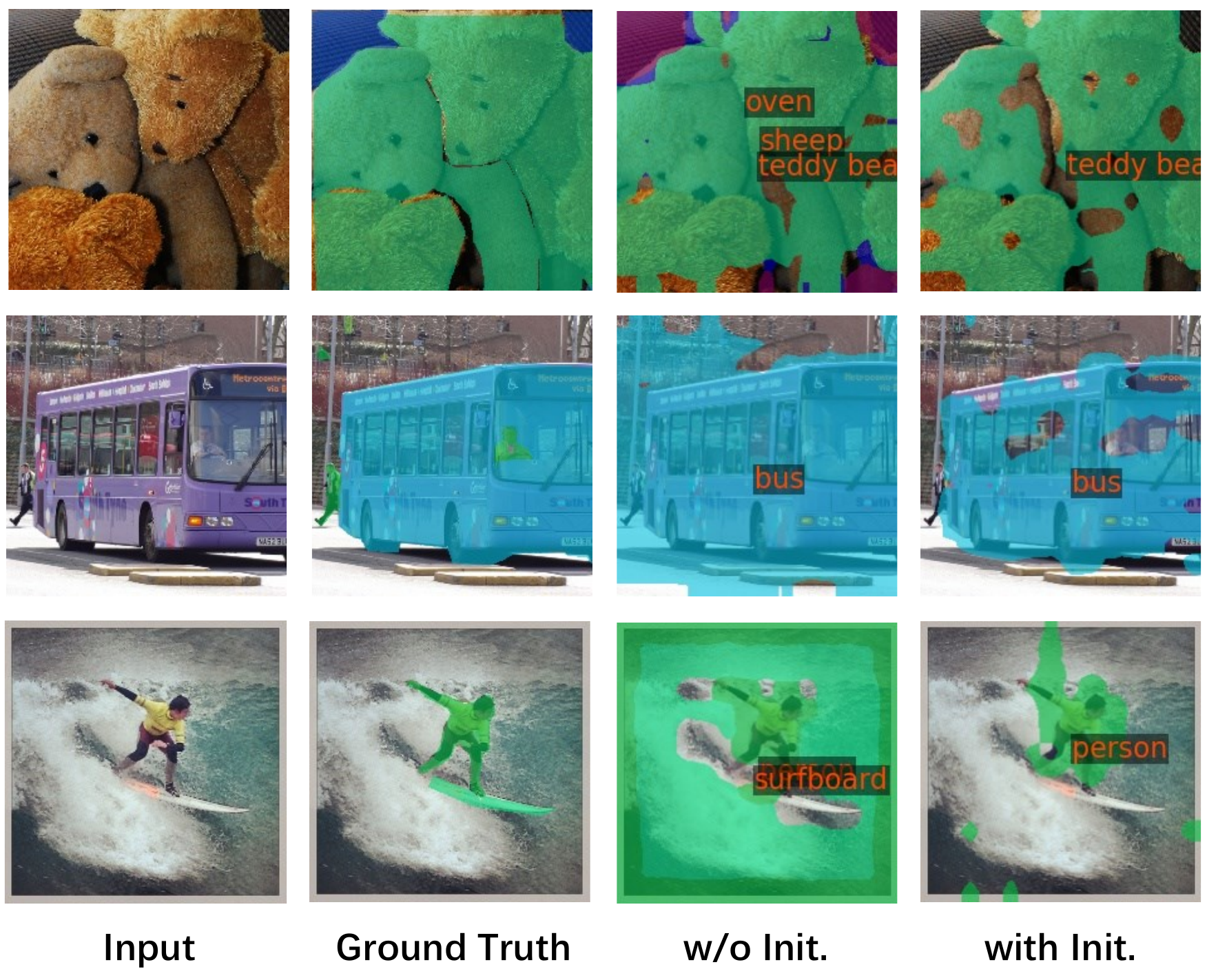}
    \caption{\textbf{Qualitative results on COCO}.}
    \label{fig:qualitative_results_coco}
\end{figure}

\subsection{Qualitative Results}
We demonstrate the qualitative results on PASCAL VOC, PASCAL Context, and COCO in Figures \ref{fig:qualitative_results_voc}-\ref{fig:qualitative_results_coco}, respectively. The results indicate that the SegCLIP can generate plausible segments and reasonable tags. Compared with the SegCLIP training from scratch, the initialized SegCLIP achieves better semantic segmentation. In Figure \ref{fig:qualitative_results_voc}, the first row implies the initialized SegCLIP could obtain better semantics, e.g., the airplane area, and the last row presents the initialized SegCLIP could obtain correct tags, e.g., dog, for the generated segments. The same conclusion could be drawn from Figure \ref{fig:qualitative_results_context} and the first two rows of Figure \ref{fig:qualitative_results_coco}. We can also observe that the single objective, multiple objects of the same class, or multiple objects from different classes can be captured by the SegCLIP. It suggests that the model training on the large scale of image-text pairs could induce the fine-grained alignment between segments and tags.

\section{Related Work}
\label{sec_related_work}
This paper is related to the vision-language pre-training and open-vocabulary semantic segmentation.

\subsection{Vision-Language pre-training}
The vision-language pre-training (VLP) is an emerging research topic with the increase of large-scale visual and linguistic pairs collected from the Internet \cite{tan2019lxmert, Chen2020UNITER, Huang2020PixelBERT, Kim2021ViLT, Li2021UNIMO, Wang2022SimVLM, Sun2019VideoBERT, Luo2020UniVL, Bain2021Frozen, Li2022BLIP, Li2022LAVENDER}.
The research directions commonly involve the design of new model architectures and pre-training objectives \cite{Gan2022VLP}. 
For the architecture, the VLP models usually contain several modules, e.g., visual encoder, text encoder, multimodal fusion encoder, or decoder. 
For the objective, the representative pre-training tasks contain the masked language model (MLM) introduced in language pre-training \cite{Devlin2019BERT}, vision-text matching (VTM),  vision-text contrastive learning (VTC), and masked vision model (MVM). Besides the model, the available datasets are the key factor in pushing the development of this research field, e.g., Conceptual Captions \cite{Sharma2018CC} and COCO \cite{Lin2014COCO} used in this work, CC12M \cite{Changpinyo2021CC12M}, YFCC \cite{Thomee2016YFCC100M}, LAION-400M \cite{Schuhmann2021LAION}, and HowTo100M \cite{Miech2019HowTo100M}.

Most vision-language pre-training models are designed for image-text or video-text related downstream tasks. Beyond that, some are mainly designed for visual tasks with text supervision. The CLIP \cite{Radford2021Learning} and ALIGN \cite{Jia2021Scaling} are two typical models trained with VTC for image classification. Further, the works from \cite{Yao2022FILIP, Zeng2022Multi} consider fine-grained alignment, and the work from \cite{Li2022Supervision} considers self-supervision within each modality plus other tasks when pre-training the model. There are also some pretrain models for object detection \cite{Gu2021Open, Zhong2022RegionCLIP, Li2022Grounded} and segmentation \cite{Wu2020PhraseCut, Ghiasi2021Scaling, Lueddecke2022Image, Rao2022DenseCLIP, Ding2022Open}.

The proposed SegCLIP is a vision-language pre-training model for segmentation. Besides our elaborate model, we also designed a reconstruct loss and a superpixel-based KL loss as our training objectives. Although the SegCLIP is capable of training with a large-scale dataset, we consider its transfer capability of reusing the existing pre-trained model, i.e., CLIP, for segmentation and reducing the cost of training resources. 

\subsection{Open-Vocabulary Semantic Segmentation}
The open-vocabulary semantic segmentation also called semantic segmentation in the wild in the literature, has been widely researched along with vision-text pretraining. Its target is to segment an image with arbitrary categories described by texts instead of fixed labeling vocabularies. 
As a pioneering work, ZS3Net \cite{Bucher2019Zero} combines a deep visual segmentation model with a generative model of class-dependent features. Such architecture allows the generation of visual samples from unseen classes via training a classifier with real visual samples from seen classes.
SPNet \cite{Xian2019Semantic} achieves that by transferring the knowledge from previously seen classes to novel classes by incorporating class-level semantic information into any network designed for semantic segmentation.

Due to the impressive zero-shot transferability of CLIP \cite{Radford2021Learning} on various downstream tasks, a research line is to leverage it for open-vocabulary semantic segmentation. DenseCLIP \cite{Rao2022DenseCLIP} is a dense prediction framework that converts the original image-text matching problem in CLIP to a pixel-text matching problem and uses the pixel-text score maps to guide the learning of dense prediction models. Unlike DenseCLIP, which needs an image decoder to generate the segments and is trained with ground-truth labels, MaskCLIP \cite{zhou2022maskclip} uses pseudo per-pixel labels generated from CLIP and self-training to achieve annotation-free segmentation. Similarly, \cite{Zabari2021Semantic} uses model interpretability to obtain pixel-level pseudo-labels from CLIP to supervise single-image segmentation methods. ZegFormer \cite{Ding2022Decoupling} decouples the zero-shot semantic segmentation into two sub-tasks, i.e., grouping the pixels into segments and classifying the segments with the CLIP. CLIPSeg \cite{Lueddecke2022Image} is a system building upon the CLIP model as a backbone and can generate image segmentations based on arbitrary prompts. OpenSeg \cite{Ghiasi2021Scaling} also involves proposal generation and segments classification as the ZegFormer, but it needs training with class agnostic mask annotations to generate mask proposals. 

Similarly, ZSSeg \cite{Xu2021A} proposes a two-stage semantic segmentation framework, with the first stage generating mask proposals and the second stage leveraging CLIP to classify the generated proposals. LSeg \cite{li2022languagedriven} uses a text encoder to provide a flexible label representation with a transformer-based image encoder trained with a contrastive objective to align pixel embeddings to the text embedding of the corresponding semantic class. OVSeg \cite{Liang2022Open} proposes to finetune CLIP on a collection of masked image regions and their corresponding text descriptions. Fusioner \cite{Ma2022Open} is a simple, lightweight cross-modality fusion module that can be used to explicitly bridge a variety of self-supervised pre-trained visual/language models for open-vocabulary semantic segmentation.

Unlike previous works, our model does not require any mask proposals or segmentation decoders. Instead, the proposed model uses a plugged semantic group module to aggregate patches as segments. Our work follows the line of GroupViT \cite{Xu2022GroupViT}, which learns segmentation masks from text supervision. However, we have different architecture compared with the GroupViT, and the proposed semantic group module makes the model capable of resuing the pre-trained weights from CLIP and training from scratch with noisy image-text pairs. Moreover, we propose two novel objectives to improve the visual representation further.

\section{Conclusion and Future Work}
\label{sec_conclusion}
This paper proposes a CLIP-based model SegCLIP for weakly-supervised semantic segmentation. The model could generate plausible segmentation results with only training on the annotation-free text-image datasets. The process does not contain the training on the labels or even the seen classes of segmentation datasets before inference, demonstrating a solid transfer character. The other advantage is the flexibility of the plugged design of the semantic group module, which brings the possibility of reusing the pre-trained CLIP weights. Besides, the proposed reconstruction loss and the superpixel-based KL loss improve performance, indicating that the image encoder's encoding capacity is essential for the semantics before giving the proper tags. In summary, the work takes a further step toward achieving fine-grained alignment, e.g., semantic segmentation in this paper, from training only on a large scale of image-text pairs.

\vspace{0.1cm}
\noindent
\textbf{Limitations} \quad The SegCLIP utilizes an interpolation operation to smooth the predicted boundaries. However, it was found that the use of regular image patches as the input to the image encoder often leads to rough predictions. The study recommends reducing the patch size to achieve smoother and more precise boundaries. 

\vspace{0.1cm}
\noindent
\textbf{Future Work} \quad To demonstrate the advantages of using smaller patch sizes, we conduct experiments on the VOC, Context, and COCO datasets using a patch size of 32, resulting in 49 patches per image. The obtained mIoU scores are 44.2\%, 22.0\%, and 21.4\% for each dataset, respectively. By comparing these scores to the mIoU scores of 52.5\%, 24.7\%, and 26.5\% achieved with a patch size of 16, which equates to 196 patches per image, it becomes evident that larger patch sizes can lead to inferior performance. Therefore, future research could concentrate on pretraining models with smaller patch sizes. 

Furthermore, additional experiments are performed using the validation sets from ADE20K \cite{Zhou2017scene} and Cityscapes \cite{Cordts2016cityscapes} to assess more complex scenes. The results show that SegCLIP achieves mIoU scores of 8.7\% and 11.0\% on ADE20K and Cityscapes, respectively, meanwhile GroupViT$_{\text{1-s}}$ achieves 4.9\% and 4.2\% mIoU on the same datasets. It is revealed that the complexity and intricacy of the scenes play a crucial role in performance, suggesting that exploring complex scenes is a promising research direction in the field of open-vocabulary segmentation.

The current superpixel generation process in SegCLIP operates as an offline module and is not trained end-to-end, making it essential to explore innovative end-to-end training techniques for this module to enhance its effectiveness. Additionally, the use of finely-divided superpixels may result in biased patches, making it necessary to consider the use of class-agnostic segmentation methods to generate improved pseudo-labels and improve performance. Furthermore, although the SegCLIP framework can utilize the pre-trained CLIP model as initialization, there is still potential for further advancements through post-pretraining on more extensive datasets such as CC12M and YFCC. These future research directions will present both challenges and exciting opportunities to enhance SegCLIP's performance to new heights. 

\section*{Acknowledgments}
This work was supported by the National Key R\&D Program of China (No. 2020AAA0108600) and the National Science Foundation of China (No. 62176221).

\bibliography{Segmentation}
\bibliographystyle{icml2023}
\end{document}